\documentclass[letterpaper, 10 pt, conference]{ieeeconf}  

\IEEEoverridecommandlockouts                              

\usepackage{amsmath} 
\usepackage{amssymb}  
\usepackage{todonotes}
\usepackage{algorithm}
\usepackage{algpseudocode}
\usepackage{subfig}
\usepackage{graphicx}

\usepackage{multirow}
\usepackage{cuted}
\usepackage[colorlinks = true, urlcolor  = blue]{hyperref}
\usepackage{cite}

\title{\LARGE \bf
A Real2Sim2Real Method for Robust Object Grasping\\with Neural Surface Reconstruction  
}

\author{Luobin Wang, Runlin Guo, Quan Vuong, Yuzhe Qin, Hao Su, Henrik Christensen
\thanks{All authors affiliated with Contextual Robotics Institute, UC San Diego.}
}

\begin{document}

\maketitle

\begin{strip}
\vspace{-1.3cm}
\centering
\begin{tabular}{cc}   
 \includegraphics[width=0.6\linewidth]{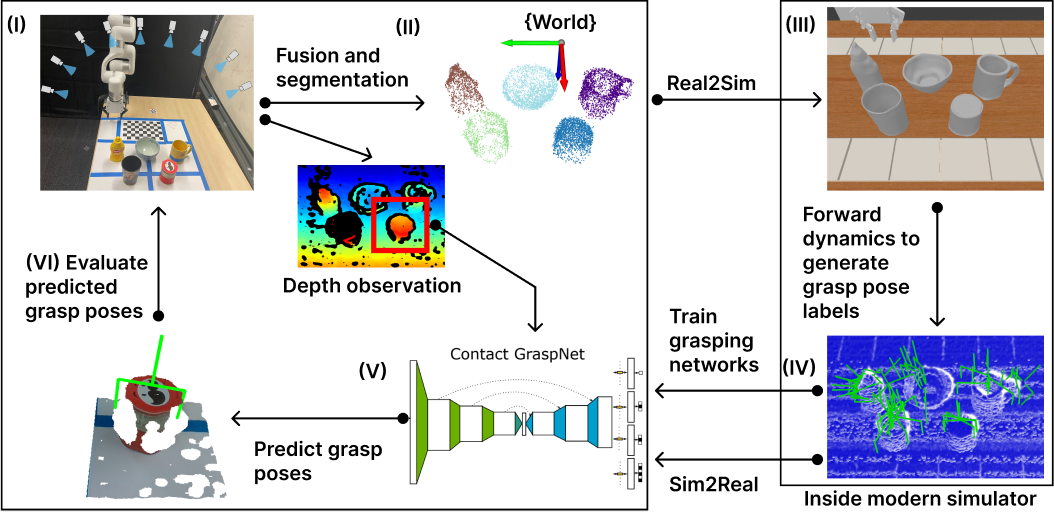} \hspace{0.2cm} &\includegraphics[width=0.37\linewidth]{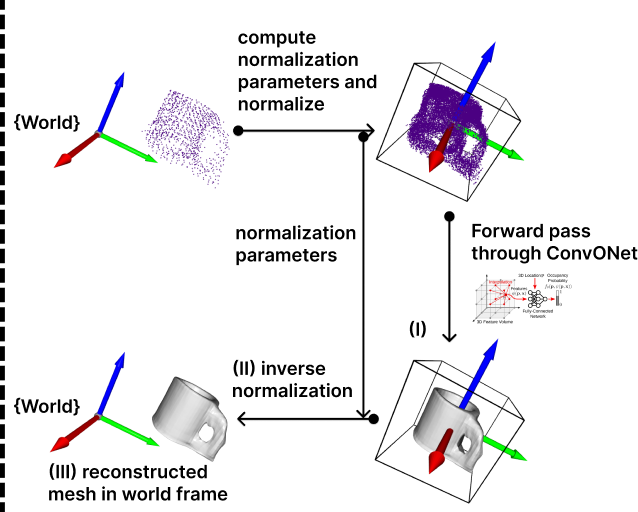}
\end{tabular}
\captionof{figure}{\textbf{Left (our pipeline):} given a real scene (I), we fuse and segment the camera observations to obtain object level point clouds (II), which we use to construct a digital replica of the real scene (III). The replica is used to generate grasp labels (IV) to obtain trained grasping networks (V). The grasp poses predicted by the trained networks are evaluated in the real scene (VI). \textbf{Right (the Real2Sim step)} illustrates how we can automatically place the reconstructed meshes in the digital replica without having to explicitly perform pose estimation. Given an object-level point cloud, we use a trained ConvONet to reconstruct the mesh (I), and then apply the inverse normalization operation (II) to obtain the mesh represented in the world frame (III). 
}
\label{fig:teaser}
\vspace{-0.4cm}
\end{strip}


\thispagestyle{empty}
\pagestyle{empty}







   




\begin{abstract}

We explore an emerging technique, geometric Real2Sim2Real, in the context of object manipulation. 
We hypothesize that recent 3D modeling methods provides a path towards building digital replicas of real-world scenes that afford physical simulation and support robust manipulation algorithm learning. Since 6 DOF grasping is one the most important primitives for all manipulation tasks, we study whether geometric Real2Sim2Real can help us train a robust grasping network with high sample efficiency. We propose to reconstruct high-quality meshes from real-world point clouds using state-of-the-art neural surface reconstruction method (the Real2Sim step). Because most simulators take meshes for fast simulation, the reconstructed meshes enable grasp pose labels generation without human efforts. The generated labels can train grasp network that performs robustly in real evaluation scenes (the Sim2Real step). In synthetic and real experiments, we show that the Real2Sim2Real pipeline performs better than baseline grasp networks trained with a $10^4\times$ larger dataset by mimicking geometric shapes of target objects in simulation. We also show that our method has better sample efficiency than training the grasping network with a retrieval-based scene reconstruction method. The benefit of the Real2Sim2Real pipeline comes from 1) decoupling scene modeling and grasp sampling into sub-problems, and 2) both sub-problems can be solved with sufficiently high quality using recent 3D learning algorithms and mesh-based physical simulation techniques. Video presentation available at \href{https://youtu.be/TkvAKLsxkSc}{this link}.

\end{abstract}
\section{INTRODUCTION}



Learning robotic manipulation skills in simulation and executing the skills in the real world, often termed \textit{Sim2Real}, has fueled many recent advances in robot manipulation \cite{DBLP:journals/corr/MahlerLNLDLOG17, 7487342, DBLP:journals/corr/abs-1709-06670, DBLP:journals/corr/abs-2011-03148, https://doi.org/10.48550/arxiv.2203.01197}. The paradigm is effective because many recent learning approaches require high volume of interaction samples (e.g., reinforcement learning \cite{zhou2022learning, agarwal2022legged} and simulation-based grasp pose auto-labeling \cite{acronym2020}), and obtaining these samples in simulators is much cheaper than in the real world. The \textit{Sim2Real} approach usually requires practitioners to build a digital replica of the real physical scene, where the robots perform the task of interest. 
To construct the digital replica in simulation, the practitioners have to manually curate the object meshes, calibrate their dynamics parameters, and place them at realistic poses. 
Even though there are existing approaches that lower the realism requirement of the digital replica, such as domain randomization and domain adaptation \cite{DBLP:journals/corr/abs-2011-03148, DBLP:journals/corr/TobinFRSZA17, DBLP:journals/corr/SadeghiL16, DBLP:journals/corr/abs-1710-06537, DBLP:journals/corr/abs-1903-11774, DBLP:journals/corr/abs-1810-02513, yu2018policy, DBLP:journals/corr/abs-1810-10093, DBLP:journals/corr/abs-1812-07252, DBLP:journals/corr/abs-1904-02750}
, these approaches still require the manual creation of 3D assets before they can be applied. Manual model creation, physical property calibration and scene construction require domain expertise and can be prohibitively costly to scale to multiple large-scale scenes with many objects \cite{habitat2, fu20213front, fu20213future, DBLP:journals/corr/abs-1904-01920}. 

Recognizing scene creation and calibration as a Sim2Real bottleneck, recent researches have attempted to automate this process and dub the problem \textit{Real2Sim2Real} \cite{Vincent2021, https://doi.org/10.48550/arxiv.1810.05687, DBLP:journals/corr/abs-2104-07662}. 
In fact, these recent researches tackle 
\textit{dynamics Real2Sim2Real}, since they focus on estimating the simulation dynamic parameters. 
We instead focus on the \textit{geometric Real2Sim2Real} challenge -- automatically constructing object geometry from camera observations, placing them at physically-plausible poses that allow for forward simulation, and demonstrating that the reconstructed scenes can train performant neural networks.  
In manipulation research, constructing object meshes and placing them into simulated scenes can be a highly manual process \cite{habitat2, AI2-THOR}. Very recently, we have witnessed breakthrough in learning-based 3D modeling. State-of-the-art neural surface reconstruction methods can convert input from 3D sensors to meshes with geometric details and have demonstrated strong cross-scene generalizability \cite{DBLP:journals/corr/abs-2003-04618, DBLP:journals/corr/abs-1901-05103, DBLP:journals/corr/abs-2106-10689, DBLP:journals/corr/abs-2003-08934}. 

\begin{figure}
\centering
\vspace{0.1cm}
{\centering
    \includegraphics[width=0.45\linewidth]{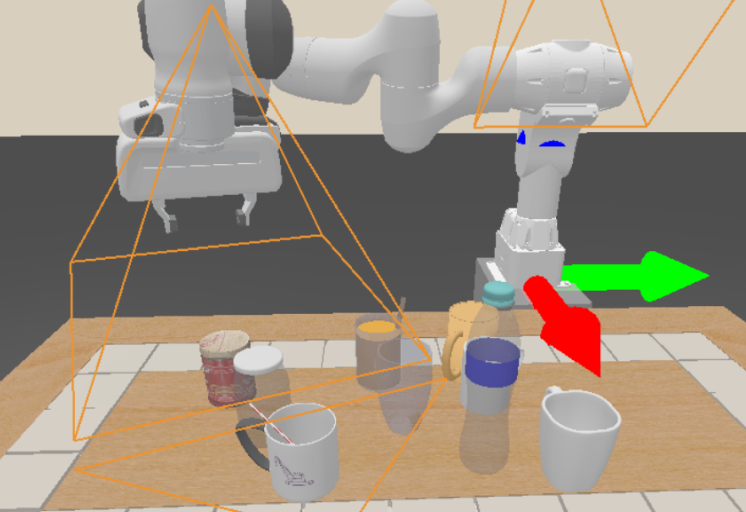}
}
\hfill
{\centering
\includegraphics[width=0.45\linewidth]{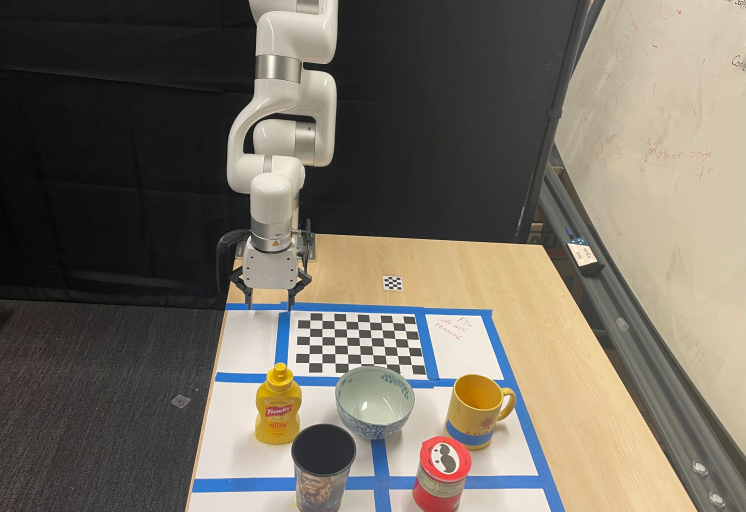}
}
\caption{
Given objects placed on a tabletop surface, the robot should pick up the objects and move them above the tabletop surface without knocking the neighboring objects over.}
\label{fig:grasping_scene_illustration} 
\vspace{-0.1cm}
\end{figure}

Therefore, it is time to ask the hypothesis whether the quality of the generated 3D meshes from state-of-the-art neural reconstruction algorithms can be used for physical simulation and 
learning of manipulation algorithms. In fact, autonomous driving researches which require coarser environment geometry have already benefited tremendously from efforts to automatically reconstruct digital clones
\cite{metasim, metasim2, virual_kitti, virual_kitti_2, Max2022}. However, manipulation research requires finer details for simulation, including precise geometric meshes and accurate physical properties.
Using 6-DoF object clutter grasping task with point cloud input as the case study, we demonstrate experimental evidences to support our hypothesis that modern neural reconstruction algorithms can provide sufficient details of objects to support simulation and learning. We choose grasping as the case study application because grasping is an essential primitives for all manipulation tasks and enabling grasping with our compact framework provides strong promises for more complex manipulation tasks. 

Our pipeline uses a performant and recent neural surface reconstruction network 
that learns to convert point clouds to detailed meshes.
To create a digital replica of the scene, we propose a simple yet effective method to place the object meshes into the scene automatically without having to explicitly perform pose estimation. We demonstrate our results in the tabletop grasping-in-clutter task and across two recent state-of-the-art grasp pose prediction networks \cite{cgn, vgn}. The grasping networks have different input modality and network architecture, providing additional evidence for the generality of our results. 

We compare our method with a recent scene reconstruction baseline \cite{ucla2021} which uses object pose estimation followed by retrieval-based method for building the digital scene replica. The results show that we can achieve higher sample efficiency if the shapes of digital replicas are closer to the real world scenario. 

To put the performance of our Real2Sim2Real pipeline into perspective, we also compare the pre-trained variant and the train-with-reconstruction variant of the best-performing grasping network we used. Although the pre-trained model used a $10^4\times$ larger training dataset, we observe similar grasping performance in simulation. We can also achieve the same robust grasping performance when adapting our compact framework to a new gripper in the real world without having to relabel the entire large dataset.
In summary, our method decouples 3D modeling and grasp pose sampling, and both sub-problems can be solved with quality and generalizability using state-of-the-art methods. 


\section{RELATED WORK}

\begin{algorithm}[t]
\vspace{0.1cm}
 \hspace*{\algorithmicindent} \textbf{Input:} depth and segmentation maps, camera poses\\
 \hspace*{\algorithmicindent} \textbf{Output:} A digital replica of the test scene
\begin{algorithmic}[1]
\State{Convert the depth maps to point clouds and fuse the point clouds}
\State{Extract object-level point clouds using the semantic segmentation maps}
\For{Each object-level point cloud}
\State{Perform point cloud outlier removal}
\State{Use \autoref{algo:obj_point_to_mesh} to reconstruct an object mesh}
\State{Place the object mesh into the simulation scene}
\EndFor
\caption{Step-by-step description of our reconstruction framework to create a digital replica of the test scenes from camera observations}
\label{algo:point_to_mesh_in_scene}
\end{algorithmic}
\end{algorithm}

\noindent\textbf{3D Reconstruction} methods are generally explicit \cite{DBLP:journals/tog/NiessnerZIS13, DBLP:journals/pami/0019H0021, DBLP:conf/eccv/WangZLFLJ18} or implicit. Implicit representation encodes the shape as continuous function~\cite{DBLP:journals/corr/abs-2003-04618, DBLP:conf/cvpr/ParkFSNL19} and can, in principle, handle more complex geometries. For Real2Sim, we utilize ConvONet~\cite{DBLP:journals/corr/abs-2003-04618}, which provides a flexible implicit representation to reconstruct 3D scenes with fine-grained details.


\noindent\textbf{Sim2Real gap} is a common problem when transferring a simulator policy to the real-world.
While existing techniques, such as domain randomization and system identification, can reduce the Sim2Real gap \cite{DBLP:journals/corr/TobinFRSZA17, DBLP:journals/corr/abs-1808-00177, DBLP:conf/icra/XieDPBG21}, they focus on generalization with respect to the low-dimensional dynamics parameters. We instead study geometric Real2Sim2Real, aligning the high-dimensional geometry to improve Sim2Real transfer.

\noindent\textbf{Geometric Real2Sim2Real} is a generative problem: given observations of a real scene, an algorithm should \textit{generate} the corresponding digital replica. Generative algorithms typically fall into two classes. Firstly, the algorithm retrieves from an external database during the generation process, an approach taken by \cite{ucla2021}. The second class of approaches relies solely on the trained generative model at test time. A representative algorithm is \cite{ditto2022}. While \cite{ucla2021, ditto2022} can reconstruct objects and scenes, they do not demonstrate that the reconstruction can be used to train networks to perform robotic manipulation tasks. The ability to use the reconstructed scenes to train networks is an important step towards solving \textit{geometric Real2Sim2Real}. Moreover, \cite{ditto2022} requires \textit{manual} placement of the reconstructed meshes into simulation. They also affix the reconstructed meshes to their supporting surface to ensure the meshes do not fall over. Such practice is not suitable for applications that require separation between the object meshes and their supporting surface, such as pick-and-place. On the other hand, \cite{ucla2021} performs reconstruction by retrieval from a database, and hence the reconstructed objects might not well approximate the real object shape. Also differently from \cite{lv2022sagcisystem}, we demonstrate that our reconstructions can enable training robust neural networks.



\section{PROBLEM STATEMENT}

Our study investigates whether state-of-the-art implicit reconstruction methods can generate digital replicas that enable automated grasp labelling and the training of robust manipulation models. Demonstrating the ability of the Real2Sim2Real approach to train robust grasping networks is crucial since 6 DoF grasping is a fundamental building block for many manipulation skills.

We focus on grasping common household objects on a typical kitchen tabletop with a parallel gripper robot arm. The objective is to grasp and lift the target object out of the structured clutter without knocking over other neighboring objects. Examples of simulated and real scenes are shown in \autoref{fig:grasping_scene_illustration}. Our synthetic object dataset includes 546 object instances from four categories (bottle, bowl, can, and mug), taken from PartNet-Mobility and ShapeNetCorev2 \cite{SAPIEN, shapenet}. Each of our $30$ simulated scenes contains $5$ to $10$ randomly chosen objects and is constructed by rejection sampling to ensure that all objects are standing upright with no collision. In the real-world scenes, we also used five objects from the same four categories. We refer to the simulated scenes and real-world scenes collectively as test scenes in our subsequent discussion.

We extensively studied the grasping problem with a hand-to-eye setting, where we position RGB-D cameras above the table and looked at the table center to capture posed depth images from the test scenes. Given posed depth images and segmentation masks of the test scenes as inputs, the reconstruction algorithm outputs posed object meshes which are used to construct a replica in a physical simulator. We then use the reconstructed scene to train grasping networks, such as Contact-GraspNet \cite{cgn} or Volumetric Grasping Network \cite{vgn}. The trained grasping network is evaluated in the original test scenes and we use the grasp success rate to measure the performance of our Real2Sim2Real pipeline. 

We demonstrate our results using test scenes in both simulation and real-world. For simulation environments, to account for realistic depth camera noises, we adapted SimKinect, which is a popular noise injection algorithm to benchmark SLAM systems \cite{handa:etal:2014, Barron:etal:2013A, Bohg:etal:2014} to add noise to the simulated depth images. Our simulator SAPIEN \cite{SAPIEN} uses PhysX \cite{physx}, a state-of-the-art physics engine, to simulate dynamic physics. To compare the grasping success of our method, we include a retrieval-based reconstruction method \cite{ucla2021} as a baseline in our simulated test scenes. The retrieval-based method belongs to a different category of reconstruction method that does not use implicit representations to recover object shapes. The retrieval-based method therefore serves as a useful baseline to understand if \textit{any} reconstruction method would lead to good reconstruction and grasping performance, or whether the particular choice of the reconstruction method leads to significant differences in grasping success.

\begin{algorithm}[t]
 \hspace*{\algorithmicindent} \textbf{Input:} $p\_WOlist \in R^{n \times 3}$, $3D$ positions of $n$ object points expressed with respect to the world frame $W$\\
 \hspace*{\algorithmicindent} \textbf{Input:} Trained Convolutional ONet, represented by $f$\\
 \hspace*{\algorithmicindent} \textbf{Output:} A mesh constructed from the input point cloud $p\_WOlist$, expressed in the world frame $mesh\_W$
\begin{algorithmic}[1]
\State{Obtain the max and min bounds of the input point cloud}
\begin{align*}
bound\_max & = np.max(p\_WOlist, axis=0) \in R^3 \\ 
bound\_min & = np.min(p\_WOlist, axis=0) \in R^3
\end{align*}
\State{Compute normalization parameters}
\begin{align*}
center & = (bound\_max + bound\_min) / 2 \\
scale & = np.max(bound\_max - bound\_min)
\end{align*}
\State{Normalize point cloud and obtain mesh}
\begin{align*}
p\_WOlist\_norm & = (p\_WOlist - center) / scale \\
mesh\_O & = MISE( f (p\_WOlist\_norm) )
\end{align*}
\State{Transform mesh to world frame}
\begin{align*}
T\_WO & = \begin{bmatrix}
scale* I  & center \\
0 & 1
\end{bmatrix} \\
mesh\_W & = mesh\_O.apply\_transform(T\_WO)
\end{align*}
\caption{Obtaining reconstructed mesh from object-level point cloud}
\label{algo:obj_point_to_mesh}
\vspace{-0.1cm}
\end{algorithmic}
\end{algorithm}

\section{Using digital replica to train networks}


We next provide description of the reconstruction framework in \autoref{sec:reconstruction_desc}, the grasping algorithm in \autoref{sec:grasping_network} and implementation details in \autoref{sec:impl_details}.

\subsection{Reconstruction framework without pose estimation}
\label{sec:reconstruction_desc}


The first step of our framework consists of converting the depth maps and segmentation masks
into object-level point clouds. That is, after this first step, each object-level point cloud only contains points that lie on the surface of the same object instance. For each object-level point cloud, we use a trained ConvONet \cite{DBLP:journals/corr/abs-2003-04618} to produce an implicit representation of the object surface. The object mesh can then be extracted by querying the learned implicit representation using the Multiresolution Isosurface Extraction procedure introduced in \cite{DBLP:journals/corr/abs-1812-03828}. 


We next describe how we obtain the object poses to place the reconstructed meshes into digital replica of the test scene. To do so, we first describe an important pre-processing step used in ConvONet. Given an object-level point cloud, represented with respect to the world frame, centering and scaling operations are applied to the object-level point cloud before inputted into ConvONet \cite{DBLP:journals/corr/abs-2003-04618}. This step transforms the object-level point cloud into a canonical frame whose origin coincides with the origin of the world frame. 
Therefore, the inverse of the centering and scaling operations can be interpreted as the object pose. We can place the reconstructed mesh into simulation without ever having to estimate the object poses explicitly. 
This is a significant advantage considering how challenging object pose estimation can be. Even though the normalization operations discussed in this paragraph are commonly used in 3D vision algorithms as a pre-processing step, we posit that the usage of the inverse normalization operations to place objects into the digital replica is a less explored and useful technique that is of independent interest.

We describe the procedure in Algorithm~\autoref{algo:obj_point_to_mesh} and visually in \autoref{fig:teaser} (Right).
The step-by-step description of our reconstruction framework is in Algorithm~\ref{algo:point_to_mesh_in_scene}. The object orientation is implicitly represented by the point cloud and hence the reconstructed object is already oriented correctly. 

\begin{figure}
\vspace{-0.3cm}
\centering
\subfloat[Without applying outlier removal, the outlier point drastically affect the centering and scaling operations, leading to incorrect reconstruction (the big blob under table).]{\label{fig:before_outlier_removal-1}
\centering
    \includegraphics[width=0.45\linewidth]{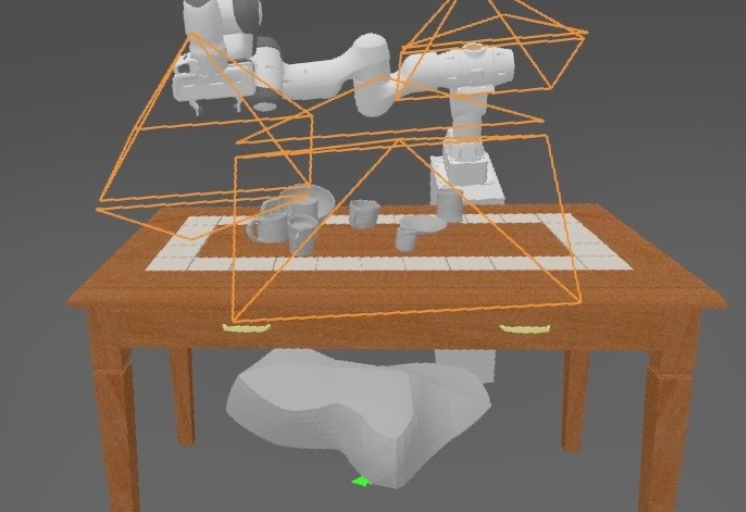}
}
\hfill
\subfloat[Reconstruction of the same scene with outlier removal applied. The previously erroneously reconstructed mesh shown in the left has more accurate reconstruction.]{\label{fig:after_outlier_removal-1}
\centering
\includegraphics[width=0.45\linewidth]{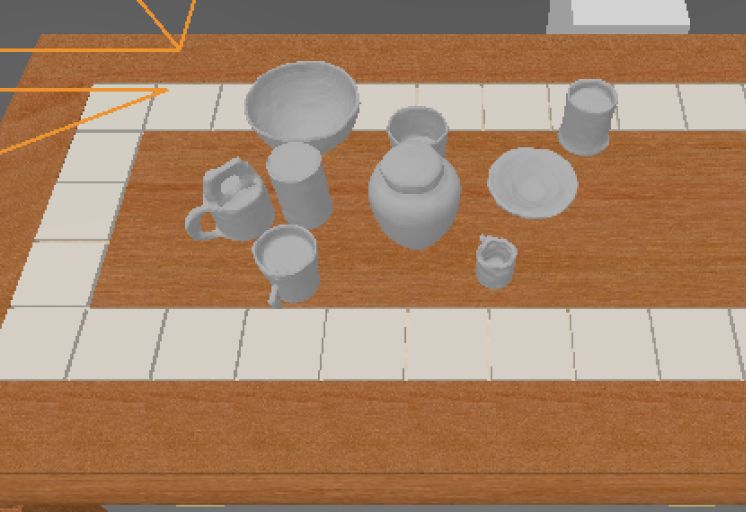}
}
\caption{
The figure illustrates the importance of applying point cloud outlier removal to the observed point cloud to improve the robustness of the reconstruction framework.
}
\label{fig:effect_of_outlier_removal}
\vspace{0cm}
\end{figure}

In addition to these steps, we apply two additional techniques to improve the reconstruction quality. First, we use statistical point cloud outlier removal \cite{zhou2018open3d} to remove outlier points in the object-level point cloud before the centering and scaling operations.
Doing so is helpful because the centering and scaling parameters are sensitive to point cloud noises. 
\autoref{fig:effect_of_outlier_removal} provides an illustrative example of the benefit of noise removal.
Second, given the object mesh constructed by the Multiresolution Isosurface Extraction (MISE) procedure, we perform approximate convex decomposition to obtain a simplified and smoother object mesh representation \cite{vhacd}, since the mesh obtained by MISE often has complex geometry, leading to slow collision detection during simulation. 

\subsection{Neural networks used for comparing grasping success}
\label{sec:grasping_network}

We evaluate the reconstruction quality using the success rates of grasping networks trained using the reconstructed scenes. This protocol differs from the standard evaluation metrics used in the computer vision community, which only measure how well the reconstructed shape approximates the ground truth shape, e.g. Chamfer distance. Using the grasping success as the evaluation metrics is important because we are interested in studying whether the reconstructed scenes can be used to train perfomant grasping networks. We next describe why Contact-GraspNet (CGN) \cite{cgn} and Volumetric Grasping Network (VGN) \cite{vgn} are reasonable grasping network choices for our use case. 

CGN takes point cloud as input and predicts for every point in the point cloud a corresponding grasp pose and the probability of grasp success. VGN accepts a TSDF representation and directly predicts for each 3D voxel the grasp pose and quality. As such, the input modalities of CGN and VGN only require depth information, and not color. Such requirement is suitable for our use case because our reconstruction framework only reconstructs the shape and not the color of the objects. We therefore only have access to the depth maps of the scenes and not color images. CGN and VGN are also recent state-of-the-art grasping networks with excellent reported performance. Their authors also release pre-trained models, e.g. a pre-trained CGN model trained using $17.7$ million simulated grasps \cite{CGN_github}. The pre-trained models serve as strong baseline for us to calibrate the performance of the models trained using our reconstructions.


\subsection{Implementation Details}
\label{sec:impl_details}

To perform grasping using CGN, given a desired pose of the gripper, we use the MPlib library \cite{MPlib} to plan a trajectory in joint space that moves the robot gripper from the current to the desired pose while avoiding collision. 
The VGN model takes as input a cubic voxel grid of dimension $40$ with voxel size $7.5mm$. The voxel grid spans a 3D cube of size $30cm$, which is not sufficient to include all the objects in our scene. To evaluate the pre-trained CGN model without changing the input voxel grid dimension, for each object in a test scene, we obtain a different cubic voxel grid of dimension $40$. The center of the voxel grid coincides with the center of the object axis aligned bounding box.
We use the pre-trained CGN models to obtain predicted grasp successes and poses for each voxel grid, after which their union constitute the predicted grasp successes and poses for the entire scene.

To train ConvONet, we generate the training data from $80$ training scenes using the same methodology we use to generate the test scenes. The objects used to generate the training and test scenes form disjoint sets.
We refer readers to {\cite{DBLP:journals/corr/abs-2003-04618, DBLP:journals/corr/abs-1812-03828, ditto2022} for details on the ConvONet training.

\section{EXPERIMENTAL RESULTS}
\label{sec:result}

\begin{table}[t]
\centering
\begin{tabular}{||c c||} 
 \hline
 Grasp Proposal Network  & Grasp Success \\
 \hline\hline
CGN trained using reconstructed scenes (Ours) & 0.93 (0.25) \\
 \hline
 CGN trained using Acronym dataset & 0.0 (0.0) \\
 \hline
\end{tabular}
\caption{The same grasping network is more robust when trained using our reconstructed scenes compared to the model trained using millions of grasps. The model trained using Acronym over-fits to geometry of the Panda robotic gripper used during training and does not generalize to the xArm gripper, which we use in our experiment. Qualitatively, the model generates grasps that are always in collision with the objects, and therefore has poor grasp success rate.}
\label{table:real_world_result}
\end{table}

Our experimental results answer the following questions:
\begin{enumerate}
    \item What are the performances of representative grasping networks trained with our pipeline against baseline models that are trained with retrieval-based reconstruction methods or pre-trained models that are trained on significantly larger datasets? 
    \item What changes do we need to make to representative grasping networks training recipe to better accommodate the proposed pipeline?
    \item Do the networks trained using the reconstructions generalize to unseen scene configurations?
\end{enumerate}

\subsection{Real-world 6-DOF grasping in clutter experiments}

The experiments using simulated scenes allow us to produce reproducible and detailed analyses in the upcoming sections. To provide evidence that our framework can generalize to real-world data, even though the reconstruction and grasping neural networks are only trained in simulation, we additionally run real-world grasping experiments using an xArm 6, illustrated in \autoref{fig:grasping_scene_illustration} (Right). \autoref{table:real_world_result} demonstrates that CGN pre-trained with Panda gripper cannot generalize to the real scene where the grasping is executed with xArm gripper. However, by training CGN with grasp pose labels generated using the xArm gripper in the reconstructions, the trained model can reach reasonable grasping performance.


\subsection{Simulated experiments comparing against pre-trained grasping models and retrieval-based reconstruction method}

\begin{table}[t]
\vspace{0.1cm}
\centering
\begin{tabular}{||c c c||} 
 \hline
 Dataset for training grasping models  & CGN & VGN \\
 \hline\hline
ShapeNet groundtruth models and scenes & 0.77 (0.16) &  0.1 (0.04)\\
 \hline
Reconstructed scenes only (ours) & 0.76 (0.15) & 0.34 (0.14)\\
 \hline
\end{tabular}
\caption{The table demonstrates the quality of the reconstructed scene. 
The CGN and VGN model trained using our reconstructed scenes performs favorably relatively to the model pre-trained using ShapeNet. Since the CGN models trained using our scenes are trained from scratch, their good performance demonstrates that the reconstructed scenes allow for training performant grasping networks.
}
\label{table:grasp_succ_cgn_implicit_vs_pretrained}
\vspace{-0.1cm}
\end{table}

\textbf{Result for CGN} \autoref{table:grasp_succ_cgn_implicit_vs_pretrained} illustrates the quality of our reconstructed scenes, in the sense that the CGN model trained using our reconstructed scenes performs well relatively to the publicly available pre-trainedd model. This is a particularly exciting result because the publicly released model was trained using $17.7$ million simulated grasps. For the results presented in \autoref{table:grasp_succ_cgn_implicit_vs_pretrained}, we train the CGN models from scratch and do not use external grasping dataset or finetune from existing network weights to avoid confounding factors when interpreting the success of the trained CGN models. We also train a different CGN model for each reconstructed scene. That is, for each reconstructed scene, we overfit a different CGN models to the reconstructed scene. As such, the good performance of the CGN models trained using the reconstructed scenes in \autoref{table:grasp_succ_cgn_implicit_vs_pretrained} can be mainly attributed to the high quality reconstruction.


\begin{table}[t]
\centering
\begin{tabular}{||c c c||} 
 \hline
 \begin{tabular}{@{}c@{}}Reconstruction\end{tabular}
    & CGN Training Epoch & Grasp Success \\
 \hline\hline
  Ours & 5000 & 0.76 (0.15) \\
  \hline
  \multirow{5}{*}{
  \begin{tabular}{@{}c@{}}Retrieval-based \\ \end{tabular}
  } & 2000 & 0.61 (0.1) \\
   & 3000 & 0.64 (0.1) \\
   & 5000 & 0.66 (0.1) \\
   & 6000 & 0.65 (0.1) \\
   & 8000 & 0.62 (0.1) \\
  \hline
\end{tabular}
\caption{We outperform a retrieval-based reconstruction baseline, in terms of the grasping success of the grasping networks trained using the reconstruction.}
\label{table:grasp_suc_implicit_vs_ucla}
\vspace{-0.1cm}
\end{table}



\textbf{Comparison to a retrieval-based method} We next compare the performance of CGN when trained using our reconstruction framework versus using a retrieval-based scene reconstruction method \cite{ucla2021}. When training using the scenes reconstructed by our method, we simply overfit the CGN model, i.e. perform gradient descent until all training losses are minimized. We find that such model selection scheme is sufficient. However, when training CGN using the scenes reconstructed by the retrieval-based method \cite{ucla2021}, we find that simply overfitting the CGN models to the reconstructions do not lead to the highest performance - possibly because the retrieved models fail to approximate the geometry of the objects in the test scenes well. We therefore evaluate the CGN models trained with different number of epochs for fair comparison. To obtain the RGBD trajectories that the retrieval-based method expects as input, we use an optimization-based exploration method to find camera trajectory that minimizes the area of occlusion \cite{7139864}. 
\autoref{table:grasp_suc_implicit_vs_ucla} illustrates that CGN has higher performance when trained using our reconstruction framework, demonstrating that not all reconstruction method leads to good performance when training neural networks. 


\textbf{Result for VGN} The VGN grasping networks trained from scratch using our reconstructions performs competitively with the pre-trained model released by VGN's authors. The results are illustrated in \autoref{table:grasp_succ_cgn_implicit_vs_pretrained}. Since we train the VGN networks from scratch and only use the reconstructed scenes for label generation, we can conclude that the good performance of the VGN models are due to the quality of the reconstructions. We note that the grasping performance of VGN and CGN are not directly comparable because we use motion planning to avoid collision when evaluating the predicted grasps for CGN, but not VGN. In the original publication, VGN does not use motion planning to avoid collision and therefore we follow the same protocol. For the pre-trained VGN model, the predicted grasps often collide with the target or neighboring objects during grasp execution. To provide more insights into the performance of VGN, we include the grasp success per object category in \autoref{table:vgn_obj_grasp_success}.

\begin{table}[t]
\centering
\begin{tabular}{||c c c c c||} 
 \hline
 Object category  & bottle & can & bowl & mug \\
 \hline 
 Grasp Success & 0.22 (0.2) & 0.31 (0.2) & 0.42 (0.3) & 0.41 (0.3) \\
\hline
\end{tabular}
\caption{Grasp success of the VGN models trained using our reconstructions on different object categories. The differences in success rate between the object categories and the high values of the standard deviation indicate that some objects are much harder for the models than others.}
\label{table:vgn_obj_grasp_success}
\vspace{0.1cm}
\end{table}

\subsection{Training Modifications to CGN and VGN}

\begin{table}[t]
\centering
\begin{tabular}{||c c||} 
 \hline
  Grasp labeling when training using reconstructions & Grasp Success \\
 \hline\hline
 Filtering using mean success rate & 0.76 (0.15) \\
 \hline
 Using one execution trial & 0.716 (0.09) \\
 \hline
\end{tabular}
\caption{The table demonstrates the importance of determining grasp success label by executing the grasp multiple times, and only label as successful grasp pose whose mean success rate is above a pre-defined threshold.}
\label{table:grasp_succ_cgn_use_mean_success_rate}
\vspace{-0.1cm}
\end{table}

Grasping networks are often trained using a large dataset of grasp labels. For example, CGN was trained using ACRONYM \cite{acronym2020}, a dataset of 17.7M simulated parallel-jaw grasps of 8872 objects. 
In contrast, we train on test scene reconstructions and then evaluate the model in the actual test scenes.
To probe the reconstruction quality, we only train on a much smaller dataset sampled from the reconstructions and thus are required to make some modifications to the training process. 

\textbf{Modifications for CGN:} CGN predicts gripper grasp poses. A motion planning algorithms converts the pose into a joint-level trajectory for the robot controller to track. As such, if the planner is stochastic, evaluating a grasp multiple times from the same initial conditions can lead to different results, depending on whether the planned joint-level trajectory leads to collision with objects in the scenes. When we generate the grasp label using the reconstructed scenes, we thus only labels as success grasps that are likely to pick up the target objects under the stochasticity of the planner. More concretely, instead of using one execution trial to evaluate the grasp \cite{acronym2020}, during grasp label generation, we evaluate a grasp pose multiple times from the same initial scene condition, and only label the pose as successful when the mean success rate is above a pre-defined threshold. \autoref{table:grasp_succ_cgn_use_mean_success_rate} illustrates quantitatively the importance of doing so. The threshold is $0.7$ in all our experiments.

\begin{figure}
\vspace{0.1cm}
\centering
{\label{fig:before_outlier_removal}
\centering
    \includegraphics[width=0.45\linewidth]{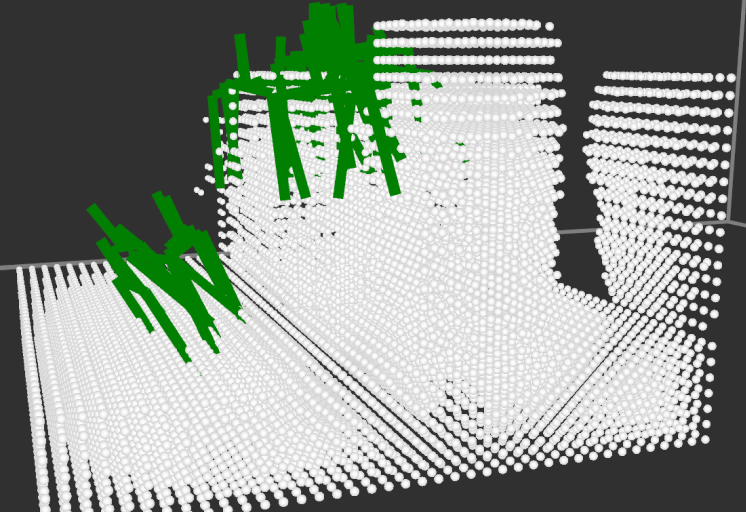}
}
\hfill
{\label{fig:after_outlier_removal}
\centering
\includegraphics[width=0.45\linewidth]{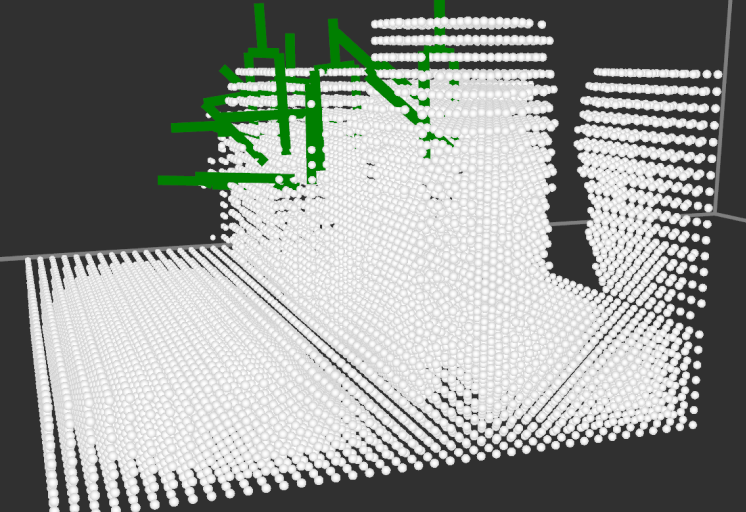}
}
\caption{\textbf{Left:} VGN assigns high success probability to voxels far away from object surface. \textbf{Right:} the network no longer assigns high success probability to these voxels when the loss is computed through all input voxels. The green grasps represent grasps with high predicted success probability.}
\label{fig:vgn_modif} 
\end{figure}

\begin{table}[t]
\vspace{0.1cm}
\centering
\begin{tabular}{||c c||} 
 \hline
 Input voxels without grasp annotations & Grasp Success\\
\hline
 Without assigning negative labels  & 34.0 \\
\hline
 With assigning negative labels  &  60.0 \\
 \hline
\end{tabular}
\caption{VGN performs better when we compute the loss through all input voxels and assign negative grasp labels to voxels without annotations. The models are trained with grasp annotation generated from the ground truth test scene.} 
\label{table:vgn_result_input_voxel}
\vspace{0.1cm}
\end{table}

\textbf{Modifications for VGN:} VGN takes as input a 3D voxel grid and predicts for each voxel a grasp pose and the corresponding success probability. Instead of only computing the loss and gradients through the voxels in which ground truth grasp label is available, we compute the loss and gradients through all voxels in the input voxel grid. For voxels where there is no grasp annotation, we label the voxel with a negative grasp label. Without our modifications, the VGN model can assign high grasp success probability in 3D region where there is no nearby voxels with grasp annotation label, as illustrated in \autoref{fig:vgn_modif} and \autoref{table:vgn_result_input_voxel}.
 
\subsection{Generalization performance to unseen scenes}
\label{sec:generalization_exp}

We previously train grasping networks using the reconstructions of the test scenes the grasping networks are evaluated in. In this section, we devise additional experiments to understand the generalization behavior of the grasping networks trained from the reconstructions as the dataset size and quality vary. Given the limited time and the amount of manual efforts to rearragne the selected scenes, we select $5$ test scenes where the CGN models (trained using the reconstructions of the test scenes the models are evaluated in) have good performance. We then vary the dataset size and quality while training CGN and evaluating the trained models in these $5$ test scenes. 

\textbf{Rearrange setting}: for each of the $5$ test scenes, we rearrange the objects in the test scene into different poses, as illustrated in \autoref{fig:rearrange_viz}. We then evaluate the CGN models trained using the reconstruction of the test scene (before rearrangement) in the re-arranged scene. In other words, the CGN models were trained on the reconstructions of the objects in the test scenes, but the arrangement of the objects has changed when evaluating the CGN models. This setting represents generalization to unseen scenes with seen objects.


\begin{figure}
\centering
{\label{fig:before_rearranging}
\centering
    \includegraphics[width=0.45\linewidth]{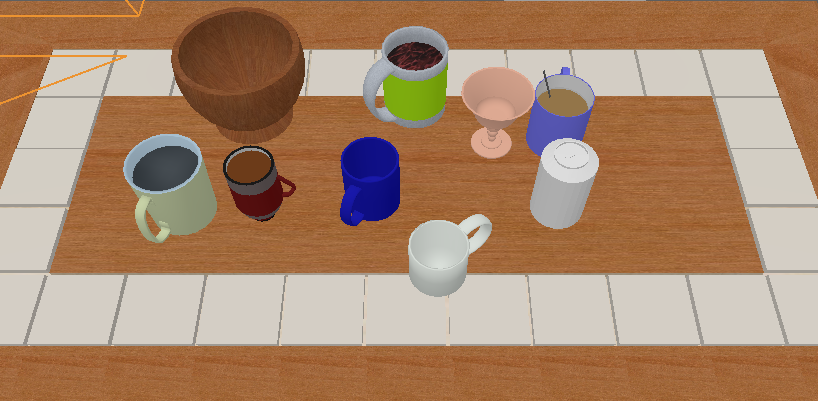}
}
\hfill
{\label{fig:after_rearranging}
\centering
\includegraphics[width=0.45\linewidth]{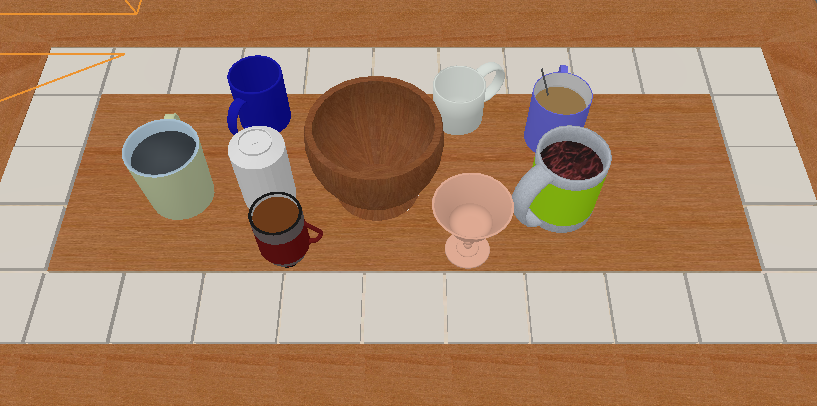}
}
\caption{\textbf{Left:} a test scene \textbf{Right:} after rearranging the objects in the left test scene, we evaluate CGN trained in the reconstruction of the \textbf{left} in the right scene.}
\label{fig:rearrange_viz}
\vspace{-0.1cm}
\end{figure}

\begin{figure}[t]
    \centering
    \includegraphics[width=0.35\textwidth]{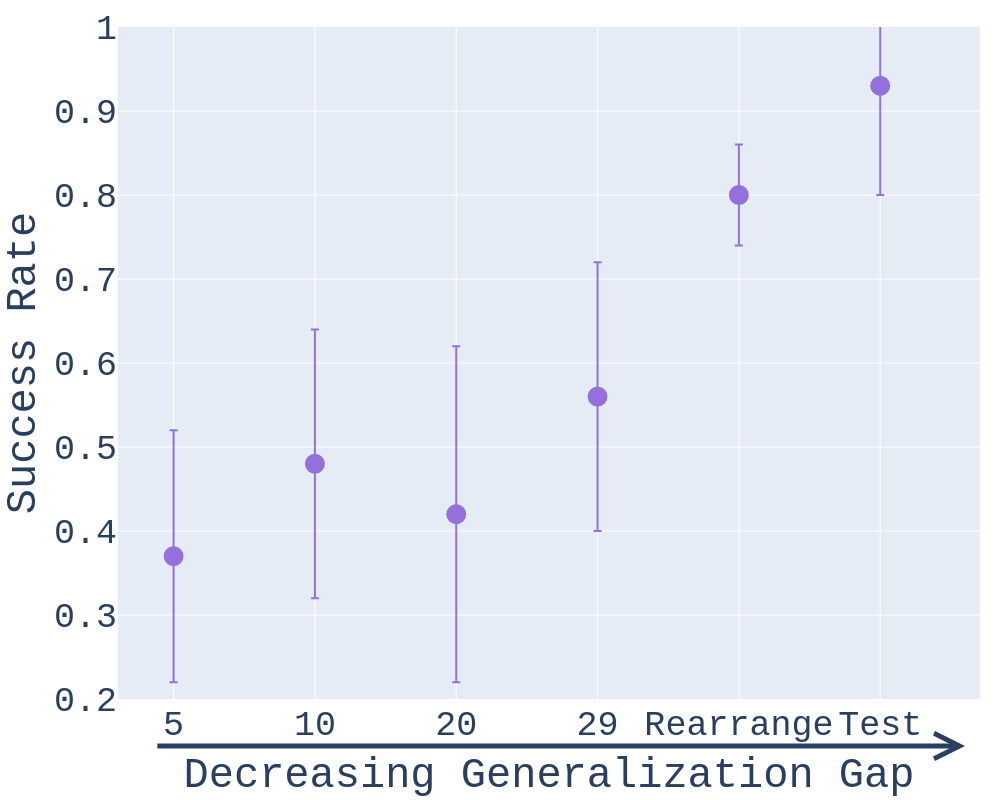}
    \caption{The performance of the Contact-GraspNet increases as a function of both the number of training reconstructed scenes
    and how close the reconstructions resemble the test scene.
    The number $5, 10, 20, 29$ indicates the number of training reconstructed scenes, holding out the test scene reconstruction.
    \textit{Rearrange} refers to the rearrange setting described in \autoref{sec:generalization_exp}. \textit{Test} refers to the setting where the CGN model is trained on the reconstructed test scene only.
    }
    \label{fig:decrease_generalization}
\vspace{-0.3cm}
\end{figure}

\textbf{Training using reconstructions of scenes different from test scenes:} in previous sections, we train one CGN models per reconstructions of the test scenes. In this experimental setting, we instead train one CGN model on reconstructions of \textit{multiple} test scenes, but evaluate the trained models in test scenes whose reconstructions are not included in the set of reconstructions used for training the CGN models.


The grasp success of the trained CGN models in these two different settings are illustrated in \autoref{fig:decrease_generalization}. The results suggest that having more reconstructed scenes improve generalization, and the trained models can generalize to rearranged scenes with a minor drop in performance.


\section{CONCLUSIONS}
We explore the geometric Real2Sim2Real pipeline on a tabletop clutter grasping task.
Our synthetic and real experiments demonstrate robust grasping results when state-of-the-art reconstruction and grasping networks are combined wisely using a modern physical simulator for bridging. We also show promising performance when adapting existing grasping networks to a new robot system following our pipeline without having to label a large dataset. 
Our work illustrates the good performance of current neural surface reconstruction methods for physical simulation and the usefulness of retraining recent grasping networks with simulated grasp sampling when deploying onto a new robot system.









\bibliographystyle{IEEEtran}
\bibliography{IEEEabrv, reference}

\end{document}